\pdfoutput=1

\documentclass[11pt]{article}

\usepackage[final]{acl}

\usepackage{times}
\usepackage{latexsym}

\usepackage[T1]{fontenc}

\usepackage[utf8]{inputenc}

\usepackage{microtype}

\usepackage{inconsolata}

\usepackage{graphicx}

\usepackage{pifont}
\usepackage{tcolorbox}
\usepackage{bbm}
\usepackage{amsmath}
\usepackage{amssymb}
\usepackage{ulem}
\usepackage{multirow}
\usepackage{booktabs}
\usepackage{caption}
\usepackage{subcaption}
\usepackage{float}

\title{Advancing MoE Efficiency: A Collaboration-Constrained Routing (\texttt{C2R}) Strategy for Better Expert Parallelism Design}
\vspace{-5cm}

\author{Mohan Zhang\thanks{Equal contribution}, Pingzhi Li$^*$, Jie Peng, Mufan Qiu, Tianlong Chen \\
University of North Carolina at Chapel Hill \\
\centering \href{https://github.com/UNITES-Lab/c2r-moe}{\raisebox{-0.1\height}{\includegraphics[height=1em]
{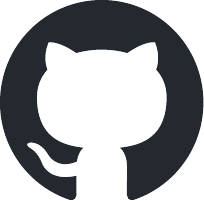} \texttt{https://github.com/UNITES-Lab/c2r-moe}}}}

\begin{document}
\maketitle
\begin{abstract}
Mixture-of-Experts (MoE) has successfully scaled up models while maintaining nearly constant computing costs. By employing a gating network to route input tokens, it selectively activates a subset of expert networks to process the corresponding token embeddings.
However, in practice, the efficiency of MoE is challenging to achieve due to two key reasons: ($1$) \textit{imbalanced expert activation}, which leads to substantial idle time during model or expert parallelism, and insufficient capacity utilization; and ($2$) \textit{massive communication overhead}, induced by numerous expert routing combinations in expert parallelism at the system level.
Previous works typically formulate it as the \textit{load imbalance} issue characterized by the gating network favoring certain experts over others or attribute it to \textit{static execution} which fails to adapt to the dynamic expert workload at runtime.
In this paper, we exploit it from a brand new perspective, \textit{i.e.}, a higher-order view and analysis of MoE routing policies: \textit{expert collaboration and specialization} --- where some experts tend to activate broadly with others (\textit{collaborative}), while others are more likely to activate only with a specific subset of experts (\textit{specialized}).
Specifically, our experiments reveal that most experts tend to be overly collaborative, 
leading to increased communication overhead from repeatedly sending tokens to different accelerators.
To this end, we ($1$) propose a novel collaboration-constrained routing (\texttt{C2R}) strategy to encourage more specialized expert groups, as well as to improve expert utilization, and ($2$) present an efficient implementation of MoE that further leverages expert specialization.
With our proposed \texttt{C2R} design, we achieve an average performance improvement of $0.51\%$ and $0.33\%$ on LLaMA-MoE and Qwen-MoE respectively across \textbf{ten} downstream NLP benchmarks, and reduce the all$2$all communication costs between GPUs, bringing an extra $20\%$-$30\%$ total running time savings on top of the existing SoTA, \textit{i.e.}~MegaBlocks.
\end{abstract}

\section{Introduction}

Scaling up the capacity of transformer models has proven to be an effective approach to enhancing model accuracy. However, as the amount of parameters increases, the immense computational and memory overheads have become a critical bottleneck~\citep{kaplan2020scaling, clark2022unified}. 
To address this challenge, the sparsely activated Mixture-of-Experts design has been introduced as a substitute for conventional Feed-Forward Networks (FFNs).
It integrates the conditional computation mechanism in the network, where only a subset of parameters is activated at runtime.
With such a property, MoE architecture has been demonstrated to successfully expand the model capacity without increasing the corresponding computational requirements~\citep{fedus2021switch, lepikhin2020gshard}, making it popular in various domains, such as language, vision, and others~\citep{lepikhin2020gshard, riquelme2021scaling, kumatani2021building}.

Despite promising potentials, the dynamic nature of MoE, stemming from well-designed routing mechanisms~\citep{pan2024dense, fedus2021switch, roller2021hash, zhou2022mixture}, also introduces new challenges during training and inference.
Extensive literature shows that the dynamic routing strategy of MoE often results in most tokens being routed to a specific subset of experts, which is termed as the load imbalance issue~\citep{lewis2021base, clark2022unified, hazimeh2021dselect, he2022fastermoe}.
This ineffective expert utilization not only leads to training instability but also hampers the full exploitation of model capacity~\citep{fedus2021switch}.
Previous works have attempted to address this issue by introducing noise to gating network or using load-balancing loss to force uniform expert activation~\citep{shazeer2017outrageously}.
However, the MoE model still suffers from massive communication overhead at the system level due to the inherently large space of expert routing combinations.
When implementing expert parallelism, each input token is dispatched to several best-fit parallel experts and the outputs are then aggregated to forward to the next layer.
Such a dispatch-aggregation process will incur tremendous redundancy if the selected experts are distributed across multiple accelerators (GPUs or nodes), making communication the inference bottleneck~\citep{he2022fastermoe, gale2023megablocks, liu2024deepseek}.

This paper delves into this problem from a novel perspective: \textit{expert collaboration and specialization}, to elucidate the routing behavior of MoE models.
This newly proposed aspect of the MoE property enables a better understanding of its inference behavior and paves a novel path for the MoE routing mechanism tailored to efficient expert parallelism design.
Specifically, we define an expert as \textit{collaborative} if it tends to collaborate (co-activate) extensively with many other experts and \textit{specialized} if it is primarily co-activated with a small group of specific experts.
Our empirical analysis reveals that most experts are overly collaborative, meaning that each expert could potentially collaborate with any other expert to handle certain tokens, which leads to the aforementioned communication overhead.
Consequently, we propose a novel collaboration-constrained routing (\texttt{C2R}) strategy at the model level to deliver a property of specialized expert groups, which provides a new chance to address this issue.
Initially, we derive the most closely collaborating expert group for each expert from our experimental analysis. 
For each token, instead of directly routing it to the top-$\mathtt{K}$ experts, we first select its top-$\mathtt{1}$ expert and then choose the remaining $\mathtt{K-1}$ experts from its corresponding collaborating expert group. 
This routing mechanism dynamically reduces the space of expert routing combinations, encouraging more specialized expert groups where only experts from the same group are collaborative.
Subsequently, we co-locate each expert within the same expert group on the same computing unit to minimize the communication overhead at the system level.
Finally, the model-system co-design achieves a Pareto optimal balance between the accuracy and efficiency of MoEs.
Evaluation across multiple benchmarks indicates the promising potential of our approach. Our key contributions are summarized below:
\begin{itemize}
    \item We present a new perspective, \textit{i.e.}, expert collaboration and specialization, to elucidate the routing behavior of MoE models and propose a novel collaboration-constrained routing (\texttt{C2R}) strategy to enhance expert utilization.
    \item Leveraging the new property of specialized expert groups, we further propose an optimized expert parallelism design at the system level to reduce communication overhead by minimizing the communication redundancy of tokens.
    \item By combining the proposed techniques at both the model and system levels, we achieve up to $24.9\%$ potential reduction in total inference wall-clock time, without compromising model accuracy. Furthermore, we even observe an average performance improvement of $0.51\%$ and $0.33\%$ on LLaMA-MoE and Qwen-MoE, respectively, across ten benchmarks.
\end{itemize}

\section{Related Works}

\paragraph{Mixture of Experts (MoE).} MoE is a distinct neural network architecture where the model's parameters are divided into multiple sub-modules, known as experts. Computations are conditionally performed by activating certain experts based on the input~\citep{jacobs1991adaptive,jordan1994hierarchical,chen1999improved,6215056}. Traditional dense MoEs are computationally heavy because they engage all experts for every input token~\citep{eigen2013learning}. Recent advancements~\citep{shazeer2017outrageously,lepikhin2020gshard,fedus2021switch} have demonstrated the effectiveness of sparsely activated MoEs (SMoEs) during both training and inference. SMoEs significantly reduce computing costs and enable language models to scale to unprecedented sizes, reaching trillions of parameters~\citep{fedus2021switch}. This efficient methodology has led to the growing adoption of SMoEs in a variety of natural language processing~\citep{shazeer2017outrageously,lepikhin2020gshard,zhou2022mixture,zhang2021moefication,zuo2022taming,jiang2021towards} and computer vision tasks~\citep{riquelme2021scaling,eigen2013learning,ahmed2016network,gross2017hard,wang2020deep,yang2019condconv,abbas2020biased,pavlitskaya2020using}.

\paragraph{Challenges in Efficient MoE Training and Inference.}
Mixture of Experts (MoE) models face significant challenges in efficient training and inference, primarily due to insufficient specialization, load imbalance, and dynamic routing strategies \citep{fedus2021switch, lepikhin2020gshard, shazeer2017outrageously}. To address these issues, researchers have focused on improving routing algorithms and enhancing communication efficiency \citep{lewis2021base, clark2022unified, nie2022hetumoe, yu2024moesys, roller2021hash, zhou2022mixture, hwang2023tutel, he2022fastermoe}. 
Specifically:
(a) MegaBlocks~\citep{gale2023megablocks} expresses MoE layer computation as block-sparse operations to accommodate imbalanced token-expert assignments. They introduced dropless-MoEs (dMoEs) and developed high-performance GPU kernels for block-sparse matrix products.
(b) Tutel~\citep{hwang2023tutel} introduced a framework that implements adaptive parallelism switching, allowing dynamic adjustment of parallelism strategies without overhead. They also developed adaptive pipelining and a 2-dimensional hierarchical All-to-All algorithm for efficient MoE computation.
(c) BASE Layer~\citep{lewis2021base} formulates MoE routing as a linear assignment problem, maximizing token-expert affinity under fully balanced allocation constraints. This method eliminates the need for load-balancing loss functions and capacity factors used in previous approaches.

Despite these advancements, existing research still focuses primarily on accelerating MoE models by adjusting the routing relationship between tokens and experts. Our method, however, goes a step further by discussing routing behavior from a higher perspective, $i.e.$, the expert collaboration and specialization.

\section{Methodology}

\begin{figure*}[t]
    \centering
    \includegraphics[width=\textwidth]{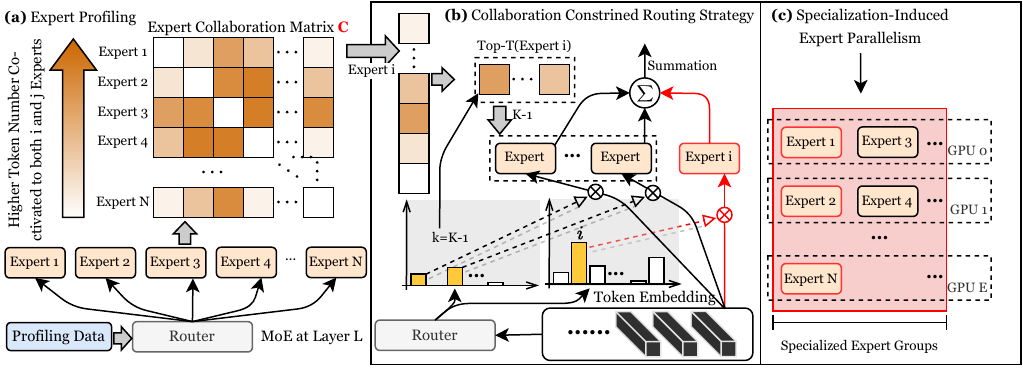}
    \vspace{-15pt}
    \caption{Overview of \texttt{C2R}. \uline{(a)} shows the process of expert profiling where we obtain the expert collaboration matrix for each layer of the MoE model; \uline{(b)} describes the mechanism of our \texttt{C2R} strategy. It first selects the $\texttt{top-}1$ expert for a given token ($\mathtt{Expert\ }i$ here) and then selects the remaining $\mathtt{K-1}$ experts from list $\texttt{Top-}\mathtt{T}(\mathtt{Expert\ }i)$; \uline{(c)} shows our efficient expert parallelism design.}
    \vspace{-15pt}
    \label{fig:pipeline}
\end{figure*}
\subsection{Preliminary}
\paragraph{Sparsely Activated MoE.} In this paper, we focus on sparsely activated MoE models, which can increase model capacity with nearly constant computational overhead.
The key components include an input-dependent sparse routing network $g(x)$ and a group of $N$ experts $\mathcal{E}=\{\mathtt{E}_i\}_{i=1}^{N}$, as shown in Figure~\ref{fig:pipeline} (a).
For each input token $x$, the routing network first calculates the probability of $x$ with respect to all experts and then dispatches it to $K$ experts with the highest probability:
\begin{equation}
    g(x)=\texttt{softmax}(\texttt{Top-}\mathtt{K}(x\cdot W_g)),
\end{equation}
where $W_g$ is the learnable parameter of $g(x)$, the ``$x\cdot W_g$'' outputs a vector of length $N$, and the $\texttt{Top-}\mathtt{K}$ function keeps only $\mathtt{K}$ largest values whose index corresponds to the selected expert. 

After the routing network, each input token $x$ is fed to its selected experts to get the output $\mathtt{E}_i(x)$.
The final output is obtained by calculating the sum of the outputs of the selected experts weighted with probabilities $g(x)$.

\paragraph{\texttt{All-to-All} Communication.} 
Training and deploying MoE models require distributed computing due to their immense computational and memory demands~\citep{dai2024deepseekmoe, fedus2021switch}.
For efficiency, both data parallelism and MoE-specific expert parallelism (a specialized form of model parallelism) are utilized~\citep{hwang2023tutel, gale2023megablocks, liu2024deepseek}.
Current MoE systems assign experts to separate computing devices~(\textit{e.g.}, GPU) in expert parallelism~\citep{fedus2021switch, lepikhin2020gshard}.
This necessitates an \texttt{all-to-all} communication to \textit{dispatch} tokens to their respective experts as determined by the routing network~\citep{gale2023megablocks}.
A second \texttt{all-to-all} communication is then required to return (\textit{combine}) the tokens to their original device in data parallelism, completing the forward pass~\citep{gale2023megablocks}. Existing frameworks, however, fail to fully exploit redundant tokens, resulting in unnecessary communication costs. Specifically, tokens that are routed to multiple experts hosted on different GPUs have to be redundantly transmitted multiple times, leading to significant inefficiencies. Our approach addresses this by minimizing such redundant token transfers, thereby reducing communication overhead and improving efficiency.

\subsection{Expert Profiling from Pre-trained Model}\label{sec:expert-profiling}
In contrast to previous studies~\citep{zoph2022st, fedus2021switch} that attempt to improve MoE efficiency from the perspective of imbalance issues~\citep{lewis2021base, clark2022unified, hazimeh2021dselect, he2022fastermoe}, in this paper, we return to the primitive goal of MoE design: different experts contain specialized knowledge, and the routing policy dynamically selects experts to process given inputs.
From this angle, we notice that the widely adopted load balancing loss~\citep{zoph2022st, fedus2021switch} in popular MoE models is insufficient~\citep{dai2024deepseekmoe, qwen_moe, zhu2024llama}, as it only targets evenly distributed expert activation during training. Such a design neglects the combinatorial aspect of the current routing mechanism, where multiple experts are activated simultaneously to collaborate on processing a given input token.
This limitation indicates that their specialization is not enough~\citep{chen2022task, xue2022one, shen2023moduleformer}, and the communication overhead remains unacceptable.
In addition, one common intuition is that the specialized knowledge in the training data is not evenly distributed, and pursuing uniformly activated experts contradicts the goal of specialization.

Therefore, based on the aforementioned observations, we aim to improve the specialization in the existing MoE model from a higher-order and novel viewpoint: collaboration and specialization among experts, specifically their co-activation patterns. 

\paragraph{Collaboration among Experts.}
Given a batch of input tokens $\mathcal{B}$, let $C_{ij}$ denote the number of times expert $i$ and expert $j$ are activated simultaneously, defined as:
\begin{equation}
C_{ij}=\sum_{x\in\mathcal{B}}\mathbbm{1}\{{g(x)_i\neq 0 \wedge g(x)_j\neq 0}\},
\end{equation}
where $g(x)_i$ represents the routing score of expert $i$ for input $x$ and $g(x)_i=0$ means the expert $i$ is not selected by token $x$. The collaboration matrix $C$ among experts is illustrated in Figure~\ref{fig:pipeline} (a). Note that we compute a collaboration matrix independently for each layer of the model because all input tokens are synchronously forwarded to the last layer.
Empirical analysis reveals that inputs with specific patterns are often handled by fixed combinations of experts. Based on this observation, we propose a metric to measure whether an expert tends to be collaborative or specialized. The collaboration degree $P_i$ of an expert is defined as the entropy of its collaboration frequency distribution with other experts:
\begin{equation}
P_i=-\sum_{\substack{j=1\\i\neq j}}^n p_{ij}\texttt{log}(p_{ij}),
\end{equation} 
where $p_{ij}=c_{ij}/\sum_{j=1}^N{c_{ij}}$ is the collaboration frequency between expert $i$ and $j$. A higher $P_i$ indicates that expert $i$ has a more uniform collaboration frequency distribution with other experts, suggesting a greater tendency for collaboration. Conversely, a lower $P_i$ implies that the expert tends to collaborate with specific experts, indicating a higher degree of specialization.
Further, we take the average of the collaboration degree $P_i$ of all experts within the same layer as the collaboration degree of that layer.
\subsection{\texttt{C2R} Strategy for Specialized Expert Groups}

Our empirical analysis of expert collaboration dynamics in pre-trained Mixture-of-Experts~(MoE) models reveals that every expert is prone to collaborate with a specific subset of experts than others. This observation forms the foundation of our proposed strategy, which aims to foster specialized expert groups by constraining the potential combinations of expert collaborations. 

\begin{table*}[t]
\begin{minipage}{0.48\textwidth}
\centering
\caption{Comparison of the performance on \textbf{reasoning tasks} and efficiency of the two evaluated network architectures using our \texttt{C2R}, Random-\texttt{C2R}, and conventional top-$\mathtt{K}$ routing strategies, respectively. Bold numbers highlight the higher accuracy or speedup ratio between our method and baselines.}
\vspace{-0.5em}
\resizebox{\textwidth}{!}{
\begin{tabular}{l|ccccc|rr}
\toprule
\midrule
\multirow{2}{*}{Methods} & \multicolumn{5}{c|}{Reasoning Tasks} & \multicolumn{2}{c}{Speedup~($\%$)} \\ \cmidrule{2-8}
& WSC & GPQA & LogiQA & PIQA & PROST & \texttt{EP=}$2$ & \texttt{EP=}$4$ \\
\midrule
\multicolumn{8}{c}{LLaMA-MoE} \\ 
\midrule
Top-$\mathtt{K}$ & $79.12$ & $\mathbf{24.37}$ & $25.04$ & $77.58$ & $25.76$ & $4.0\uparrow$ & $3.0\uparrow$ \\ 
Random-\texttt{C2R} & $78.39$ & $24.15$ & $\mathbf{25.19}$ & $76.28$ & $\mathbf{26.18}$ & $4.2\uparrow$ & $4.8\uparrow$ \\ 
\texttt{C2R} & $\mathbf{80.22}$ & $24.11$ & $\mathbf{25.19}$ & $\mathbf{77.86}$ & $25.75$ & $\mathbf{4.5\uparrow}$ & $\mathbf{13.5\uparrow}$ \\
\midrule
\multicolumn{8}{c}{Qwen-MoE} \\
\midrule
Top-$\mathtt{K}$ & $\mathbf{77.66}$ & $30.34$ & $30.88$ & $78.84$ & $30.59$ & $15.8\uparrow$ & $18.9\uparrow$  \\
Random-\texttt{C2R} & $76.19$ & $28.09$ & $30.26$ & $78.35$ & $29.70$ & $16.2\uparrow$ & $21.0\uparrow$  \\
\texttt{C2R} & $76.92$ & $\mathbf{30.41}$ & $\mathbf{31.64}$ & $\mathbf{78.94}$ & $\mathbf{31.05}$ & $\mathbf{17.6}\uparrow$ & $\mathbf{24.9}\uparrow$ \\
\midrule
\bottomrule
\end{tabular}}
\vspace{-1em}
\label{tab:table1}
\end{minipage}
\hfill
\begin{minipage}{0.46\textwidth}
\centering
\caption{Comparison of the performance on \textbf{NLU tasks} and efficiency of the two evaluated network architectures using our \texttt{C2R}, Random-\texttt{C2R}, and conventional top-$\mathtt{K}$ routing strategies, respectively. Bold numbers highlight the higher accuracy or speedup ratio between our method and baselines.}
\vspace{-0.5em}
\resizebox{\textwidth}{!}{\begin{tabular}{l|ccccc|rr}
\toprule
\midrule
\multirow{2}{*}{Methods} & \multicolumn{5}{c|}{NLU Tasks} & \multicolumn{2}{c}{Speedup~($\%$)} \\ \cmidrule{2-8}
& RACE & SciQ & RTE & BoolQ & COPA & \texttt{EP=}$2$ & \texttt{EP=}$4$ \\
\midrule
\multicolumn{8}{c}{LLaMA-MoE} \\ 
\midrule
Top-$\mathtt{K}$ & $39.52$ & $88.90$ & $49.10$ & $72.63$ & $84.00$ & $4.0\uparrow$ & $3.0\uparrow$ \\ 
Random-\texttt{C2R} & $39.52$ & $89.20$ & $50.18$ & $70.92$ & $80.00$ & $4.2\uparrow$ & $4.8\uparrow$ \\ 
\texttt{C2R} & $\mathbf{39.90}$ & $\mathbf{89.60}$ & $\mathbf{50.54}$ & $\mathbf{72.91}$ & $\mathbf{85.00}$ & $\mathbf{4.5\uparrow}$ & $\mathbf{13.5\uparrow}$ \\
\midrule
\multicolumn{8}{c}{Qwen-MoE} \\
\midrule
Top-$\mathtt{K}$ & $\mathbf{39.14}$ & $\mathbf{94.70}$ & $72.92$ & $80.03$ & $80.00$ & $15.8\uparrow$ & $18.9\uparrow$ \\
Random-\texttt{C2R} & $38.66$ & $94.60$ & $67.87$ & $75.17$ & $\mathbf{84.00}$ & $16.2\uparrow$ & $21.0\uparrow$  \\
\texttt{C2R} & $\mathbf{39.14}$ & $\mathbf{94.70}$ & $\mathbf{73.29}$ & $\mathbf{80.24}$ & $82.00$ & $\mathbf{17.6}\uparrow$ & $\mathbf{24.9}\uparrow$ \\
\midrule 
\bottomrule
\end{tabular}}
\vspace{-1em}
\label{tab:table2}
\end{minipage}
\end{table*}

\paragraph{\texttt{C2R} Mechanism}
To summarize, our proposed approach restricts expert collaboration combinations by leveraging empirical observations of collaboration patterns, thereby encouraging the formation of more specialized and efficient expert groups.
\ding{172} To implement the \texttt{C2R} strategy, we start with analyzing expert collaboration patterns in pre-trained models using a representative corpus that simulates realistic token distributions. 
Specifically, we take tokens from the corpus as input to the MoE model and feed them forward to the final layer. For each layer $l$ of the model, we first obtain an expert collaboration matrix $C_l \in \mathbb{N}^{N \times N}$ as described in Section~\ref{sec:expert-profiling} where $N$ is the number of experts in one layer and $C_l(i,j)$ denote the number of tokens routed to both expert $i$ and expert $j$ simultaneously. Then, we sort each row of this matrix and select the top $\mathtt{T}$ indices $\mathcal{I}_\mathtt{T} \in \mathbb{N}^{N \times T}$, resulting in a list of the $\mathtt{T}$ most frequently collaborating experts for each expert, denoted as:
\begin{equation}
    \texttt{Top-}\mathtt{T}(\mathtt{E}_i)=\{ e_j \in \mathcal{E} \,|\, j \in \mathcal{I}_\mathtt{T}[i] \}
\end{equation}
where $\mathtt{T} \in [1, N]$ is the hyperparameter controlling the degree of collaboration, and will be analyzed in detail in Section~\ref{sec:expert-collaboration-analysis}.\ding{173} Building upon this analysis, we propose the \texttt{C2R} strategy. 
For each token, instead of directly routing it to the corresponding top-$\mathtt{K}$ experts, we first select its $\texttt{top-}1$ expert $\mathtt{E}_i$. Then we restrict the selection of the remaining $\mathtt{K-1}$ experts to the list $\texttt{Top-}\mathtt{T}(\mathtt{E}_i)$ identified as having the most frequent collaborations with expert $\mathtt{E}_i$. Note that the selection is still based on routing scores but is constrained to the list $\texttt{Top-}\mathtt{T}(\mathtt{E}_i)$. This approach substantially reduces the potential combinations of experts involved in the routing process, fostering more specialized expert groups while preserving the flexibility needed for dynamic routing.

\subsection{Specialization-Induced Zero-Redundancy \texttt{All-to-All}}

MoE layers often underutilize GPUs due to sequential \texttt{all-to-all} communications and feed-forward layers for token dispatching and combining. The \texttt{all-to-all} communication typically consumes over $30\%$ of runtime~\citep{hwang2023tutel}, with this proportion increasing as the number of GPUs grows, leading to inefficient MoE expert parallelism.

Our proposed \texttt{C2R} routing strategy offers additional opportunities to optimize communication overhead. By recognizing that certain expert collaborations occur more frequently, we can co-locate closely collaborating experts on the same computational units (\textit{e.g.}, GPUs). For tokens routed to multiple experts residing on the same device, we employ a novel design that sends only a single copy of the token to the shared device, where it is subsequently replicated locally for processing by each assigned expert. This approach significantly reduces inter-device communication, thereby alleviating \texttt{all-to-all} communication bottlenecks. As expert collaboration patterns stabilize through specialization, this optimized strategy can yield substantial reductions in communication costs, ultimately improving the model's overall efficiency.

\section{Experiments}

\subsection{Implementation Details} 

\paragraph{Network Architecture and Baselines.} We select two representative open-source MoE models, including LLaMA-MoE~\citep{zhu2024llama} and Qwen-MoE~\citep{qwen_moe}. LLaMA-MoE has $32$ transformer layers, hidden size $4096$, $32$ attention heads, and $8$ experts per MoE layer with top-$2$ routing. Qwen1.5-MoE has $24$ layers, hidden size $2048$, $16$ attention heads, and $60$ experts per MoE layer with top-$4$ routing, plus a shared expert. We implement our method by replacing the original top-$\mathtt{K}$ routing with our \texttt{C2R} strategy in both models, leaving Qwen-MoE's shared experts unchanged. For baselines, we use model's original routing policy as our baseline, and replace the expert profiling process of \texttt{C2R} with random initialization as another baseline. Each expert in LLaMA-MoE is an MLP with an intermediate dimension of $1376$ and input/output dimension of $4096$, while in Qwen1.5-MoE, experts have an intermediate dimension of $1408$ and input/output dimension of $2048$, with the shared expert's intermediate dimension being $5632$.
\paragraph{Datasets and Benchmarks.} We conduct supervised fine-tuning of the two selected models under the original routing policy and ours, respectively. For LLaMA-MoE, we follow the script provided by its official repository to fine-tune on the Deita-6K dataset~\citep{liu2024what}. For Qwen-MoE, we choose the popular LIMA instruction tuning dataset~\citep{zhou2024lima}. We examine the superior performance of our proposed routing strategy on popular benchmarks and potential inference speedup. Specifically, $10$ benchmarks across two types of downstream tasks are examined in this paper, including natural language understanding (RACE~\citep{lai-etal-2017-race}, SciQ~\citep{SciQ}, RTE~\citep{wang2019rte}, BoolQ~\citep{clark2019boolq}, COPA~\citep{roemmele2011choice}) and reasoning (WSC~\citep{levesque2012winograd}, GPQA~\citep{rein2023gpqa}, LogiQA~\citep{liu2020logiqa}, PIQA~\citep{bisk2020piqa}, PROST~\citep{aroca2021prost}). We use WikiText~\citep{merity2016pointersentinelmixturemodels} for expert parallelism profiling in efficiency evaluation.

\paragraph{Training Configuration.} For LLaMA-MoE, we follow the official training settings to do full-parameter SFT, using AdamW optimizer~\citep{yao2021adahessian} and the learning rate is set to $2\times 10^{-5}$ with a warm-up ratio of $0.03$ and cosine scheduler. We use a total batch size of $16$ with gradient accumulation steps as $8$. The max length of the input sequence is set to $2048$. We train the model for $2$ epochs without freezing the gate. For Qwen-MoE, we basically follow the above settings, except for using: ($1$) a warm-up step of $10$ instead of a warm-up ratio of $0.03$, ($2$) a total batch size of $32$ with gradient accumulation steps set to $2$, and ZeRO-3 Offload to avoid out-of-memory (OOM) issues. We set hyperparameters $\mathtt{T}$ as $5$ and $30$ for LLaMA-MoE and Qwen-MoE, respectively.

\begin{figure*}[ht!]
    \centering 
    \includegraphics[width=0.8\linewidth]{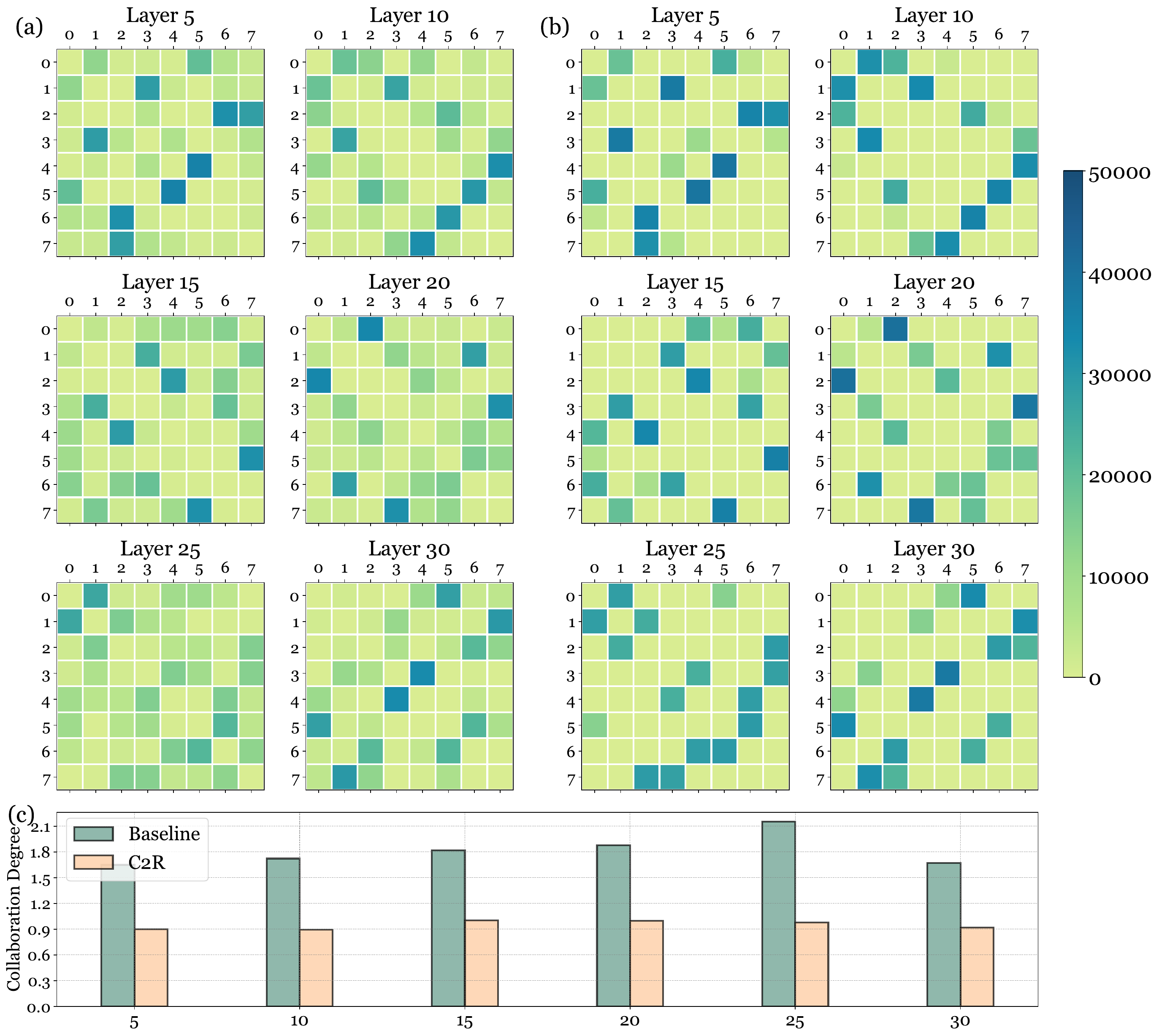}
    \vspace{-5pt}
    \caption{\small Visualization of expert collaboration matrix in several intermediate layers of LLaMA-MoE after SFT. (a): Results with conventional top-$\mathtt{K}$ routing strategy. (b): Results with our \texttt{C2R} strategy ($\mathtt{T}=2$). (c): The average collaboration degree comparison between Baseline and our \texttt{C2R} strategy. A darker pixel in (a) and (b) indicates a higher number of tokens routed simultaneously to the corresponding experts (indexed by row and column) within the given layer, which means these two experts collaborate more frequently. Note that many pixels in (b) have a value of 0, meaning that the corresponding two experts will never be selected simultaneously, while most of the pixels in (a) have a light color indicating a non-0 value. (c) demonstrates that experts in our model exhibit a higher degree of specialization.}
    \vspace{-15pt}
    \label{fig:collaboration}
\end{figure*}

\paragraph{Expert Parallelism Speedup Estimation.}
To quantify the potential speedup of our zero-redundancy \texttt{all-to-all} in \texttt{C2R}, we conducted a comprehensive analysis using a calibration dataset of $64$ random $2048$-token segments from WikiText. Our evaluation process comprised three key steps: \uline{First}, we implemented expert parallelism based on the MegaBlocks framework~\citep{gale2023megablocks} and used PyTorch Profiler\footnote{https://pytorch.org/docs/stable/profiler.html} to measure the wall-clock time proportion $\mathtt{P}^{\texttt{EP}}$ of \texttt{all-to-all} communication relative to total inference time at various expert parallelism degrees $\texttt{EP}$. \uline{Second}, we calculated the token redundancy $\mathtt{r}^{\texttt{EP}}$ for each GPU during \texttt{all-to-all} communication across different $\texttt{EP}$ values. \uline{Finally}, we derived the estimated speedup at each $\texttt{EP}$ by computing $\mathtt{P}^{\texttt{EP}} \times \mathtt{r}^{\texttt{EP}}$. This methodology allowed us to quantify the efficiency gains of our approach, considering both communication overhead and potential reductions in data transfer across various scales of expert parallelism. We evaluate the speedup across $\mathtt{EP}$ in $\{2, 4\}$.

\subsection{Superior Performance of Our Method}
We select LLaMA-MoE and Qwen-MoE models to train on Deita-6K and LIMA datasets, respectively, and the evaluation results are summarized in Table~\ref{tab:table1} and Table~\ref{tab:table2}. The following observations can be drawn: ($1$) Our \texttt{C2R} strategy demonstrates superior performance compared to the baseline routing strategies. Specifically, LLaMA-MoE with our approach achieves an average performance improvement of $0.26\%$ on the reasoning tasks and $0.76\%$ on the natural language understanding tasks, respectively, compared to the top-$\mathtt{K}$ baseline. Similarly, Qwen-MoE demonstrates improvements of $0.13\%$ and $0.52\%$ on these tasks, respectively. This validates the effectiveness of our proposed method. ($2$) Our method shows consistent performance benefits on both MoE Architectures. Specifically, we obtain a total average performance improvement of $0.51\%$ on LLaMA-MoE and $0.33\%$ on Qwen-MoE, respectively, across all datasets. This verifies the generalization of our proposed method. ($3$) Our speedup results demonstrate the significant efficiency gains of our \texttt{C2R} framework. At $\texttt{EP}=4$, Llama-MoE with \texttt{C2R} achieves around $10\%$ more speedup ratio compared to the baselines. This improvement highlights the effectiveness of our approach in optimizing communication in \texttt{all-to-all} and reducing overhead in MoE, potentially enabling more efficient scaling.

\subsection{Expert Collaboration Analysis on LLaMA-MoE}\label{sec:expert-collaboration-analysis}
\vspace{-5pt}
In this part, we take LLaMA-MoE as an example to conduct an in-depth analysis of expert collaboration under our \texttt{C2R} strategy compared to the conventional top-$\mathtt{K}$ routing strategy. Specifically, we randomly sampled a total of $200$ context sentences from different domains of the LongBench dataset~\citep{bai2023longbench}, truncating each sentence to $1024$ tokens. Overall, we used these $0.2M$ tokens to simulate token distribution in real-world scenarios and fed them into the model to derive the expert collaboration matrices $C_l$ for each layer $l$. We visualized these matrices using heatmaps as illustrated in Figure~\ref{fig:collaboration} (a) and (b). Here, we show the results of $6$ intermediate layers selected at an interval of five layers for both settings after fine-tuning. Figure~\ref{fig:collaboration} (c) shows the calculated values of collaboration degree introduced in Section~\ref{sec:expert-profiling}. From analysis, our findings are as follows: ($1$) Under the conventional top-$\mathtt{K}$ routing strategy, as shown in Figure~\ref{fig:collaboration} (a), the space of expert routing combinations is considerably large. While the collaborative tendency among experts is not uniform, where each expert tends to be co-activated more with certain experts than with others, every expert has the opportunity to collaborate with any other expert. This results in significant communication overhead when implementing expert parallelism, as many tokens tend to activate experts distributed across different accelerators. ($2$) In contrast, under our \texttt{C2R} strategy, as shown in Figure~\ref{fig:collaboration} (b), each expert is predominantly co-activated with a small group of specific experts. This greatly reduces the routing space and thus contributes to specialized expert groups, which is also supported by the calculated collaboration degree shown in Figure~\ref{fig:collaboration} (c). With this property, we can easily distribute experts from the same group on the same accelerator, thereby reducing communication costs.

\subsection{Pareto Optimal Balance between Collaboration and Specialization}

\begin{figure*}[ht]
    \centering 
    \includegraphics[width=1.01\linewidth]{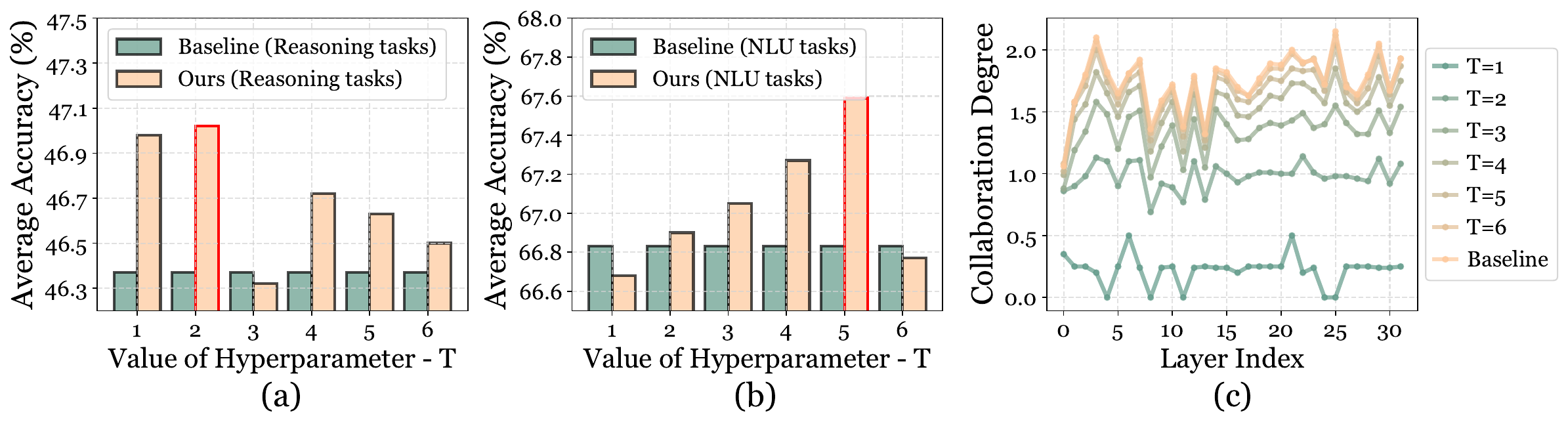}
    \vspace{-25pt}
    \caption{\small Performance and collaboration degree comparison of LLaMA-MoE. (a) and (b) respectively show the performance comparison between our \texttt{C2R} strategy (\textit{Ours}) and conventional top-$\mathtt{K}$ routing strategy (\textit{Baseline}) on two downstream tasks, namely Reasoning tasks and NLU tasks, with hyperparameter $\mathtt{T}$ varying from 1 to 6. (c) shows the collaboration degree comparison between \textit{Baseline} and \textit{Ours} under different values of hyperparameter $\mathtt{T}$ in different layers of the model. Note that since the LLaMA-MoE model we use is to select 2 out of 8 experts per layer, our method degenerates to a conventional top-$\mathtt{K}$ routing strategy (i.e., the baseline) when $\mathtt{T}=7$, so we omit this case.}
    \vspace{-15pt}
    \label{fig:hypermeterT}
\end{figure*}
In this section, we aim to answer the research question of "how much collaboration is needed." To validate our claim that there exists a Pareto optimal balance between collaboration and specialization, we vary the hyperparameter $\mathtt{T}$ from $1$ to $6$ and examine the average performance on the aforementioned benchmarks as well as the collaboration degree in each layer of the model. A smaller $\mathtt{T}$ indicates that each expert collaborates with only a few specific experts, leading to greater specialization, whereas a larger $\mathtt{T}$ implies more collaboration. This is demonstrated experimentally in Figure~\ref{fig:hypermeterT} (c). As shown in Figure~\ref{fig:hypermeterT} (a) and (b), we observe that as $\mathtt{T}$ increases, the trend of model performance initially improves and then declines. This observation supports the proposition that it is possible to find a Pareto optimal point where model performance is maximized. We also notice that there is a minimum value of model accuracy in reasoning tasks when $\mathtt{T}=3$, which we treat as noise that does not affect the overall trend. Taking communication overhead into account, the value of $\mathtt{T}$ effectively controls the trade-off between model performance and efficiency. Specifically, in some cases, we might accept a slight performance drop in exchange for more specialized expert groups (\textit{i.e.}, a smaller $\mathtt{T}$), allowing each accelerator to host an entire expert group and thereby reducing communication costs.

\subsection{Study on the Expert Parallelism Degree}
There is a trade-off between the expert parallelism degree $\texttt{EP}$ and inference speed in \texttt{C2R}. As the expert parallelism degree $\texttt{EP}$ increases: ($1$)~the vanilla \texttt{all-to-all} communication~(\textit{i.e.} \textit{w/o} zero-redundancy design) cost tends to increase due to more GPUs needed; ($2$)~the redundancy tends to decrease due to the fewer experts on each GPU. Therefore, the final inference speed with zero-redundancy \texttt{all-to-all} needs further investigation. Therefore, we evaluate the efficiency performance on Qwen-MoE across \texttt{EP} in \{$2$,$3$,$4$,$5$,$6$\}, as shown in Table~\ref{tab:ep-scaling}. The highest potential speedup rate can be achieved at $\texttt{EP}=5$, balancing the two opposing trends when equipped with zero-redundancy \texttt{all-to-all}.

\begin{table}[ht]
\caption{Efficiency performance analysis of Qwen-MoE model across expert parallelism (\texttt{EP}) dimensions from $2$ to $6$. As \texttt{EP} increases, communication redundancy decreases, but total \texttt{all-to-all} time rises. The highest potential speedup rate can be achieved at $\texttt{EP}=5$, balancing these opposing trends.}
\vspace{-5pt}
\resizebox{0.95\linewidth}{!}{
\label{tab:ep-scaling}
\begin{tabular}{c|cc|r}
\toprule
\midrule
\texttt{EP} & Redundancy & \texttt{All-to-All} Time & Speedup \\
\midrule
$2$  & $\mathbf{58.3\%}$           & $30.1\%$      & $17.6\%$   \\
$3$  & $47.6\%$           & $40.7\%$      & $19.4\%$   \\
$4$  & $40.2\%$           & $61.9\%$      & $24.9\%$   \\
$5$  & $38.4\%$           & $76.3\%$      & $\mathbf{29.3\%}$   \\
$6$  & $32.9\%$           & $\mathbf{77.2\%}$      & $25.4\%$  \\
\midrule
\bottomrule
\end{tabular}}
\end{table}

\section{Conclusion}

In this paper, we present a brand new perspective for analyzing MoE routing behavior, namely expert collaboration and specialization.
We design a novel collaboration-constrained routing (\texttt{C2R}) strategy to improve expert utilization, which further delivers a property of specialized expert groups.
Based on such characteristics, we propose an efficient expert parallelism design to reduce communication overhead at runtime.
Extensive experiments on two representative MoE models across multiple downstream benchmarks exhibit a consistent performance improvement, demonstrating the effectiveness of our approach.
Additional runtime analysis shows significant reductions in total running time savings, underscoring our design as a promising direction for addressing both training and inference efficiency challenges.

\section{Limitations}

Our study reveals an intriguing yet unexplored observation (Figure~\ref{fig:hypermeterT}): different downstream tasks (\textit{i.e.}, Reasoning and NLU) require varying degrees of expert collaboration for optimal performance. This finding suggests two promising directions for future research: ($1$) developing task-specific routing strategies that achieve a Pareto-optimal balance between collaboration and specialization, potentially yielding significant improvements in task-specific performance; and ($2$) implementing our zero-redundancy \texttt{all-to-all} approach efficiently on real-world GPU architectures. These avenues for future work address both the theoretical foundations and practical implementations of MoE models, potentially leading to more efficient and task-adaptive LLMs.

\section*{Acknowledgment}
Pingzhi Li and Tianlong Chen are supported by NIH OT2OD038045-01 and the UNC SDSS Seed Grant.

\bibliography{custom}




\end{document}